\documentclass[10pt,twocolumn,letterpaper]{article}

\usepackage{cvpr}              

\usepackage{graphicx}
\usepackage{amsmath}
\usepackage{amssymb}
\usepackage{booktabs}
\usepackage{pifont}
\usepackage{makecell}
\usepackage{multirow}
\usepackage[noEnd]{algpseudocodex}
\usepackage{enumitem}
\usepackage[percent]{overpic}%
\usepackage{verbatim}
\usepackage{bbm}
\usepackage{footnote}
\usepackage{tabularx}
\usepackage{arydshln} 
\usepackage{makecell}

\usepackage{subcaption} 
\usepackage{rotating} 
\usepackage{tikz} 

\newcommand{\darkyellow}[1]{{\textcolor[RGB]{245,194,66}{#1}}}

\newcommand{\red}[1]{{\color{red}#1}}

\colorlet{colorLow}{darkgray!70}    

\usepackage[accsupp]{axessibility}

\definecolor{cvprblue}{rgb}{0.21,0.49,0.74}
\usepackage[pagebackref,breaklinks,colorlinks,allcolors=cvprblue]{hyperref}

\def\paperID{10428} 
\def\confName{CVPR}
\def\confYear{2025}

\newcommand\myvspace{\vspace{0mm}}
\newcommand{\ours}{PanoGS}

\title{PanoGS: Gaussian-based Panoptic Segmentation for 3D Open Vocabulary \\ Scene Understanding}

\author{Hongjia Zhai$^{1}$ \quad Hai Li$^{2}$ \quad Zhenzhe Li$^{1}$ \quad Xiaokun Pan$^{1}$ \quad Yijia He$^{2}$ \quad Guofeng Zhang$^{1}$\footnotemark[2] \\
$^{1}$State Key Lab of CAD \& CG, Zhejiang University \quad $^{2}$RayNeo
}

\begin{document}
\twocolumn[{%
    \renewcommand\twocolumn[1][]{#1}%
    \maketitle
    \centering
    \vspace{-0.5cm}
\includegraphics[width=0.99\linewidth]{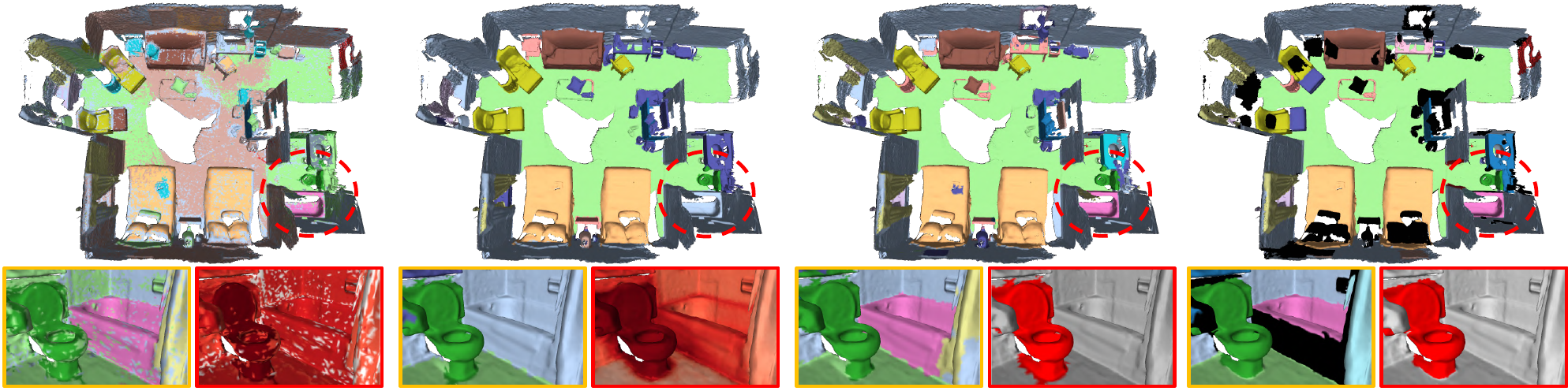}
\vspace{1em}
 \begin{minipage}{0.24\linewidth}
 \centering
 \small
 OpenGaussian~\cite{wu2024opengaussian}
 \end{minipage}
 \hfill
  \begin{minipage}{0.25\linewidth}
 \centering
 \small
 OpenScene~\cite{Peng2023OpenScene}
 \end{minipage}
 \hfill
  \begin{minipage}{0.25\linewidth}
 \centering
 \small
 \textbf{Ours}
 \end{minipage}
 \hfill
  \begin{minipage}{0.24\linewidth}
 \centering
 \small
 Ground Truth
 \end{minipage}
 \vspace{-5mm}
\captionof{figure}{\textbf{Open Vocabulary 3D Panoptic Scene Understanding}. Visualization of open-vocabulary semantic segmentation (\darkyellow{yellow boxes}) and object query with text \textit{toilet} (\red{red boxes}). 
Our {\ours} can achieve more accurate segmentation results and generate 3D instance-level results for open-vocabulary text queries, unlike previous methods that generate heatmaps between scene features and text queries.}
\label{fig:teaser}    
\vspace{4mm}
}]

\renewcommand{\thefootnote}{\fnsymbol{footnote}}
\footnotetext[2]{Corresponding author.}


\begin{abstract}
Recently, 3D Gaussian Splatting (3DGS) has shown encouraging performance for open vocabulary scene understanding tasks. However, previous methods cannot distinguish 3D instance-level information, which usually predicts a heatmap between the scene feature and text query. 
In this paper, we propose PanoGS, a novel and effective 3D panoptic open vocabulary scene understanding approach. Technically, to learn accurate 3D language features that can scale to large indoor scenarios, we adopt the pyramid tri-plane to model the latent continuous parametric feature space and use a 3D feature decoder to regress the multi-view fused 2D feature cloud. 
Besides, we propose language-guided graph cuts that synergistically leverage reconstructed geometry and learned language cues to group 3D Gaussian primitives into a set of super-primitives.
To obtain 3D consistent instance, we perform graph clustering based segmentation with SAM-guided edge affinity computation between different super-primitives. Extensive experiments on widely used datasets show better or more competitive performance on 3D panoptic open vocabulary scene understanding. 
Project page: \href{https://zju3dv.github.io/panogs}{https://zju3dv.github.io/panogs}.
\end{abstract}

\section{Introduction}
\label{sec:intro}
3D scene understanding is a critical problem in computer vision that enables humans or intelligent agents to comprehensively understand the 3D scenes and facilitate the downstream applications~\cite{huang23vlmaps,hugs,nis-slam,neuraloc,vox-fusion,ming2024benchmarking}.
They usually incorporate vision-language models (VLMs)~\cite{clip,Lseg,dino} for a fine-grained and holistic understanding of the environment.

Recently, Neural Radiance Fields (NeRF)~\cite{mildenhall:2020:nerf} and 3D Gaussian Splatting (3DGS)~\cite{kerbl3Dgaussians} have rapidly gained much research attention for novel view synthesis. 
Due to the fast rendering ability and explicit scene representation of 3DGS, it has been widely integrated into  reconstruction~\cite{chen2024pgsr,Huang2DGS2024,yu2024gof}, generation~\cite{tang2023dreamgaussian,yi2023gaussiandreamer}, and understanding~\cite{zuo2024fmgs,zhou2024feature_3dgs,gaussian_grouping}.
While the domain of combining 3DGS with scene understanding has recently made several progress~\cite{qin2024langsplat,shi2024_gs_language_embed,zuo2024fmgs,guo2024semantic_gaussian,bhalgat2024n2f2}, these approaches are primarily designed for 2D pixel-level semantic segmentation with rendered 2D language feature maps, which cannot distinguish the different objects with the same semantics in the 3D space.

Although previous 3DGS-based approaches~\cite{qin2024langsplat,shi2024_gs_language_embed,wu2024opengaussian,bhalgat2024n2f2} have achieved impressive performance that combines VLMs with 3DGS for open vocabulary scene understanding, there also exist the following limitations that prevent 3DGS-based approaches for panoptic open vocabulary scene understanding: 
\textit{1) inaccurate 3D language feature learning}. 
The discrete features attached to the Gaussian primitive can affect the inherent smoothness of language features for semantically similar objects~\cite{qin2024langsplat,zhou2024feature_3dgs,shi2024_gs_language_embed}.
The alpha-blending technique accumulates the 3D discrete features of primitives based on the opacity weight, which leads to a domain gap between 2D and 3D feature space~\cite{splatloc,wu2024opengaussian}.
And the 2D feature compression~\cite{qin2024langsplat} and feature distillation~\cite{shi2024_gs_language_embed,zuo2024fmgs} will inevitably damage the distinguishing ability of learned language features.
\textit{2) unable to recognize 3D instance-level information}. 
Previous methods~\cite{Peng2023OpenScene,wu2024opengaussian} usually predict a heatmap of similarity between the language feature of the 3D scene and the text query.
These approaches may lead to inconsistent instance segmentation results and can not distinguish the multi-instance objects of the same semantics. 
However, instance-level information is essential for 3D panoptic scene understanding.

To address the aforementioned two limitations, we propose {\ours}, a novel and effective 3DGS-based approach for 3D panoptic open vocabulary scene understanding.
Firstly, to enhance the representation ability and spatial smoothness of our learned language feature, we use a pyramid tri-plane to model the latent continuous parametric feature space of the 3D scene. 
We use the 2D multi-view fused feature cloud and confidence to perform distillation of learned language features in 3D space, not 2D rendered space, which can avoid the domain gap between 2D and 3D feature spaces caused by alpha-blending. 
In addition, to obtain 3D instance information, we formulate 3D instance segmentation as a graph clustering problem. 
To build the 3D scene graph, we use our language-guided graph cuts algorithm to group Gaussian primitives into geometrically and semantically consistent graph vertices, that is super-primitives.
Besides, we use the 2D segmentation model SAM~\cite{sam} to obtain 2D mask labels and construct the affinity between 3D super-primitives based on the mult-view consistency of 2D mask label distribution.
Finally, a progressive clustering strategy is used to obtain globally consistent instance information.
Overall, the technical contributions of our approach are summarized as follows:
\begin{itemize}
    \item{We propose {\ours}, the first 3DGS-based approach for 3D panoptic open vocabulary scene understanding.}
    \item{We learn an inherent smooth and accurate 3D language feature field based on our latent pyramid tri-plane, which is optimized by 2D fused feature cloud and confidence.}
    \item{We design an effective graph clustering based segmentation algorithm to synergistically leverage geometric and semantic cues to obtain consistent 3D instances.}
    \item{We conduct extensive experiments on commonly used datasets to demonstrate the 3D panoptic segmentation performance of our approach.}
\end{itemize}

\section{Related Work}
\label{sec:related_work}

\noindent\textbf{Panoptic Segmentation.}
The 2D panoptic segmentation task was first proposed by Kirillov~\etal~\cite{kirillov2019panoptic}. 
While many methods~\cite{cheng2020panoptic,cheng2021mask2former,porzi2019seamless} focus on improving the reasoning ability of CNN models to understand individual images, there is still a gap in 3D panoptic scene understanding due to the lack of 3D training data.
To enhance the 3D panoptic segmentation, some works aim to lift 2D panoptic predictions into 3D scene space, with different scene representations, such as point cloud~\cite{gasperini2021panoster,zhou2021panoptic}, voxels~\cite{narita2019panopticfusion}, 3DGS~\cite{wang2024plgs}, and implicit representation~\cite{siddiqui2023panoptic-lifting,fu2022panopticnerf,kundu2022panopticneuralfiled,zhu2025pcf-lifting}.
However, those works are rather limited to close-set panoptic segmentation and can not recognize the objects of unseen classes.

\myvspace\noindent\textbf{3D Gaussian Splatting.}
Recently, 3DGS~\cite{kerbl3Dgaussians} has demonstrated remarkable advancements in many tasks of 3D computer vision~\cite{splatloc,Huang2DGS2024,tang2023dreamgaussian,chen2024pgsr}.
Compared to previous NeRF-based approaches~\cite{mildenhall:2020:nerf,imtooth,Vox-Surf}, 3DGS represents the scene with a set of anisotropic 3D Gaussian primitives explicitly.
To enforce geometric consistency, some studies aim to control the shape of the primitives~\cite{Huang2DGS2024}, use unbiased depth rendering~\cite{chen2024pgsr}, and introduce geometric regularization during the optimization process, such as monocular depth~\cite{monogs,yan2024gs_slam}, and normals~\cite{turkulainen2024dn_splatter,xiang2024gaussianroom}.
Besides, some works extend 3DGS to model dynamic scenes with deformable fields~\cite{yang2024deformable_gaussian} and explicit motion estimation~\cite{shape-of-motion,stearns2024dynamic_marbles}.
Meanwhile, some work equips 3DGS with scene understanding by extending each primitive with learnable language embeddings for open-vocabulary 3D queries.
LangSplat~\cite{qin2024langsplat} uses an auto-encoder to compress the dimension of language features. N2F2~\cite{bhalgat2024n2f2} uses a tri-plane~\cite{chan2022eg3d} as the additional feature encoding to reduce parameters.

\myvspace\noindent\textbf{3D Open Vocabulary Scene Understanding.}
Inspired by the successful 2D visual-language models (VLMs)~\cite{clip,dino,Lseg}, some approaches aim to learn 3D consistent feature fields to model the semantic property with explicit 3D structures~\cite{Peng2023OpenScene} (\eg, point clouds).
Besides, some researchers try to perform scene understanding with neural implicit representations~\cite{mildenhall:2020:nerf,kerbl3Dgaussians}. 
To achieve this, LERF~\cite{lerf} and N3F~\cite{n3f} are early exploratory works, which optimize an additional field branch to align the feature space with VLMs (\eg, CLIP~\cite{clip}, DINO~\cite{dino}).
Recent efforts~\cite{qin2024langsplat,shi2024_gs_language_embed,zuo2024fmgs,zhou2024feature_3dgs} have combined 3DGS with 2D scene understanding techniques due to its advantage of explicit representation.
However, they cannot distinguish the different objects with the same semantics, which is not suitable for panoptic scene understanding.

\section{Method}
\label{sec:method}
\begin{figure*}[ht!]
\centering
\includegraphics[width=\linewidth]{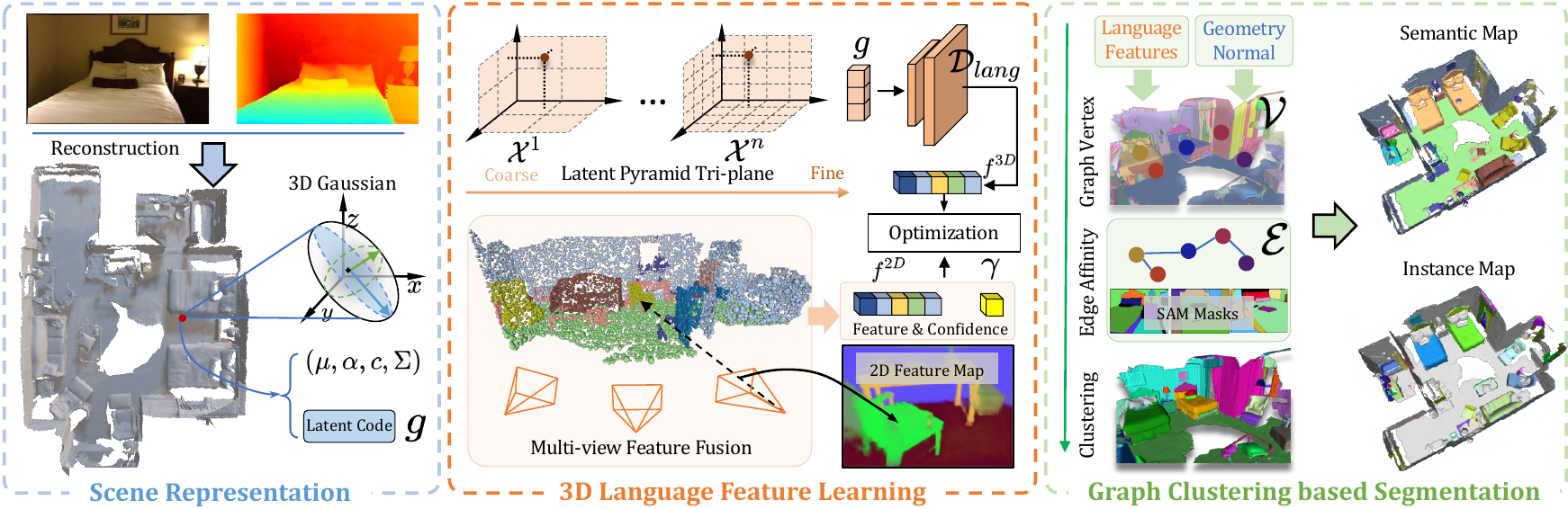} 
\caption{Overview of our approach. (a) Given posed RGB-D images, we reconstruct the scene with 3D Gaussian primitives, and each primitive is associated with additional latent language code $g$ generated from a latent continuous pyramid tri-plane feature space. (b) After the geometry reconstruction, we obtain 2D fused primitive-level features and confidences via back projection, which is used for efficient 3D language feature regression and latent pyramid tri-plane and 3D decoder optimization. (c) We perform a language-guided graph cuts algorithm to construct super-primitive and use the 2D instance mask generated by SAM~\cite{sam} to conduct progressive graph clustering.}
\label{fig:pipline}
\end{figure*}

As shown in~\cref{fig:pipline}, given multi-view posed images $\{I_i, D_i\}_{i=1}^m$, we can perform 3D open vocabulary scene understanding.
To achieve this goal, we propose an effective 3D language feature field learning module, which adopts a pyramid tri-plane to model the latent continuous parametric feature space and regress the language feature from fused feature cloud and confidence (\cref{subsec:feature_learning}).
Besides, we apply a graph clustering based segmentation algorithm to obtain 3D consistent instances based on the scene graph reconstructed with geometry and language cues (\cref{subsec:graph_clustering}).

\subsection{Scene Representation}
\label{subsec:scene_representation}
Benefiting from the efficiency of 3DGS~\cite{kerbl3Dgaussians}, we take it as our scene representation for 3D panoptic scene understanding.
Following the fast differentiable rasterization~\cite{max1995optical}, we can render the 3D scene properties to the 2D image plane.
\begin{equation}
    \hat{A} =\sum_{i} a_i \cdot \alpha_i \cdot \prod_{j=1}^{i-1}(1-\alpha_j),
    \label{eq:rendering}
\end{equation}
where $\hat{A}$ are the rendered 2D scene information (\eg, color, depth). $a_i$ and $\alpha_i$ denote the 3D property and opacity contribution of $i$-th Gaussian primitive, and $\prod_{j=1}^{i-1}(1-\alpha_j)$ is the accumulated transmittance.

Following previous works~\cite{monogs,kerbl3Dgaussians}, the appearance and geometry loss functions are used for optimization:
\begin{equation}
\mathcal{L}_{recon} = w_{1} \cdot \mathcal{L}_{c}(\hat{I},I) + w_{2} \cdot \mathcal{L}_{d}(\hat{D},D),
\label{eq:geo_color}
\end{equation}
where \{$w_{i}$\} are the weights for each optimization component and $\mathcal{L}_c$ and $\mathcal{L}_d$ are the L1 loss terms between rendered color/depth and input color/depth.

Similar to previous 3DGS-based scene understanding works~\cite{zhou2024feature_3dgs,shi2024_gs_language_embed,qin2024langsplat,zuo2024fmgs,wu2024opengaussian}, we additionally attach a latent low-dimensional code $g$ for each Gaussian primitive.
Different from them, to reduce the memory requirement and spatial noise introduced by the discrete representation, we don't explicitly save the individual features but model it via a latent continuous pyramid tri-plane feature space.
This design allows us to learn better scene language feature fields for panoptic 3D scene understanding.

\subsection{3D Language Feature Learning}
\label{subsec:feature_learning}
As an explicit modeling method, 3DGS can not store high-dimensional language features for each primitive. 
Previous approaches~\cite{shi2024_gs_language_embed,qin2024langsplat,zuo2024fmgs} apply quantization or compression to reduce the dimensions.
However, these operations inevitably reduce the accuracy and distinguishability of the learned language features.
To learn accurate 3D language features, we regress the feature via decoding the latent language code sampled from the 3D pyramid tri-plane.

\myvspace\noindent\textbf{Latent Pyramid Tri-plane.}
As shown in the middle part of~\cref{fig:pipline}, we adopt a pyramid tri-plane to model the latent continuous parametric feature space of the 3D scene. 
Compared with the previous discrete feature of each Gaussian primitive, our method can directly regress the language features of the 3D scene, which is not affected by the bias introduced by the alpha-blending as pointed out in~\cite{wu2024opengaussian,splatloc}.
Specifically, given a 3D position $\mu$, we first query its multi-resolution latent language code via the following equation:
\begin{equation}
    g(\mu) = \sum_{i}^n \{\mathcal{T}(\mu,\mathcal{X}^{i}_{xy}),\mathcal{T}(\mu,\mathcal{X}^{i}_{yz}),\mathcal{T}(\mu,\mathcal{X}^{i}_{xz})\},
\end{equation}
where $\mathcal{T}(\cdot)$ is the trilinear interpolation operation, and $\mathcal{X}_{xy}^{i}$, $\mathcal{X}_{yz}^{i}$, $\mathcal{X}_{xz}^{i}$ represent the decomposed feature planes of $i$-th resolution level in the pyramid.

Due to memory constraints, the dimension of the latent code $g(\mu)$ is usually much less than the original language feature.
So, to obtain the high-dimensional language feature for open vocabulary scene understanding, we use the 3D language feature decoder $\mathcal{D}_{lang}$, which transforms the low-dimensional latent code into high-dimensional language feature with the following equation:
\begin{equation}
    f^{3D}(\mu) = \mathcal{D}_{lang}(g(\mu)),
\end{equation}
where $f^{3D}(\mu)\in \mathbb{R}^{D_l}$ is the decoded high-dimensional language and $D_l$ is the dimension of the language feature.

\myvspace\noindent\textbf{Multi-view Feature Fusion.}
After obtaining the reconstructed Gaussian primitives, we can project primitives into multi-view 2D images.
We use LSeg~\cite{Lseg} to extract visual-language feature maps for RGB images.
So, for a 3D Gaussian primitive $p_i$, we can obtain its multi-view 2D feature vectors $\{f_1, \cdots, f_m\}$ from $m$ 2D feature maps.
To obtain its fused primitive-level feature $f_{i}^{2D}$, we adopt the weighted average on the multi-view 2D feature vectors according to the occlusion and observation of the 3DGS primitive, $f_{i}^{2D} = \Phi( \{f_1, \cdots, f_m\})$ and $\Phi(\cdot)$ is the pooling operation.
Additionally, the fused 3D primitive-level features may have different confidence due to multi-view inconsistency and occlusion.
To measure this, we compute the confidence value $\gamma_i^{2D}$ for $i$-th primitive $p_i$ as follows:
\begin{equation}
    \gamma_i^{2D} = \frac{\textit{Obs}(p_i)}{\sum_{D_l}\textit{Var}(\{f_1,f_2,\cdots,f_m\})},
    \label{eq:confidence}
\end{equation}
where $\textit{Obs}(p_i)$ count the normalized number of valid observations of primitive $p_i$, $\textit{Var}(\cdot)$ denote the variance of the observed multi-view language features $\{f_1,f_2,\cdots,f_m\}$, and $D_l$ is the dimension of the observed language feature $f_i$.

So, for the reconstructed scene with $N$ Gaussian primitives, we can obtain the primitive-level 2D fused feature cloud $\{f_i^{2D}\}_{i=1}^N$ and confidence $\{\gamma_i^{2D}\}_{i=1}^N$.

\myvspace\noindent\textbf{Language Feature Distillation.}
With the fused features cloud and confidence, we can optimize our 3D latent pyramid tri-plane $\{\mathcal{X}_{xy}^i,\mathcal{X}_{yz}^i,\mathcal{X}_{xz}^i\}$ and 3D language feature decoder $\mathcal{D}_{lang}$ to learn accurate feature representation.
Specifically, the latent codes $g$ from the pyramid tri-plane are assigned to 3D Gaussian primitives, and the 3D language features of primitives are decoded from their latent codes. So, for $i$-th primitive, its 2D fused language feature $f_i^{2D}$ and 3D learned feature $f^{3D}_i$, we can use the following equation for optimization:
\begin{equation}
    \mathcal{L}_{feat} = \sum_{i}^N \gamma_i^{2D} \cdot |1 - \cos(\mathcal{D}_{lang}(g_i), f_i^{2D})|,
    \label{eq:feat_loss}
\end{equation}
where $\cos(\cdot)$ denotes the cosine similarity function.

\subsection{Graph Clustering based Segmentation}
\label{subsec:graph_clustering}
Previous methods~\cite{siddiqui2023panoptic-lifting,zhu2025pcf-lifting} lift 2D information to 3D space for feature learning, which may lead to 3D multi-view inconsistencies.
Different from them, we directly cluster the reconstructed 3D Gaussian primitives into several disjoint subsets, where each subset can represent a class-agnostic instance in the 3D scene.
So, to achieve this goal, we formulate this problem as a graph clustering task and construct the 3D scene graph $\mathcal{G}=(\{\mathcal{V}_i\}, \{\mathcal{E}_{ij}\})$ based on our reconstructed Gaussian primitives and learned 3D language feature in~\cref{subsec:feature_learning}.
In the following, we elaborate on the details of how to construct the graph vertex $\{\mathcal{V}_i\}$, edge affinity $\{\mathcal{E}_i\}$, and progressive graph clustering.

\myvspace\noindent\textbf{Graph Vertex Construction.}
Viewing each 3D Gaussian primitive as a vertex and building a fully connected edge weight graph between all vertexes is impractical for solving graph clustering problems due to indoor scenes usually containing millions of 3D Gaussian primitives.
Previous works~\cite{yin2024sai3d,guo2022sam_graph,yan2024maskclustering} use normal information and graph cuts algorithm~\cite{felzenszwalb2004efficient_graph_cut} to group 3D points into a set of superpoints.
However, only using the geometric properties can lead to over-segmentation or under-segmentation. 
Benefiting from our language feature learned in~\cref{subsec:feature_learning}, we can simultaneously take the local geometry information and global semantic information into consideration for grouping individual Gaussian primitives into a set of super-primitives. 

To obtain geometrically and semantically consistent super-primitives $\{\mathcal{V}_i\}$, we perform the language-guided graph cuts with our language feature during the merge process.
Additionally, to access the global semantic property of super-primitives, we retrieve the language feature of the current super-primitive, which is updated during the merge process.
In language-guided graph cuts, to judge whether two super-primitives merge into a new vertex, we adopt the following criteria:
\begin{equation}
    \Delta = \mathbbm{1}\left((n_i \cdot n_j) > \lambda_n \right) \cdot \mathbbm{1}\left((f_i^{3D} \cdot f_j^{3D}) > \lambda_f \right),
\end{equation}
where $\mathbbm{1}(\cdot)$ is the indication function that equals 1 when the condition is satisfied, $n_i$ and $f_i^{3D}$ are the normal and language features within the current super-primitive. $\lambda_n$ and $\lambda_f$ are the threshold parameters at this iteration in our language-guided graph cuts process.

With our language-guided graph cuts algorithm, we can obtain geometrically and semantically consistent super-primitives rather than only using geometry information.
After traversing all reconstructed Gaussian primitives, we view each super-primitive as a vertex $\mathcal{V}_i$ in the graph $\mathcal{G}$, which is represented by the disjoint sets of super-primitives.

\begin{figure}[ht!]
\centering
 \includegraphics[width=\linewidth]{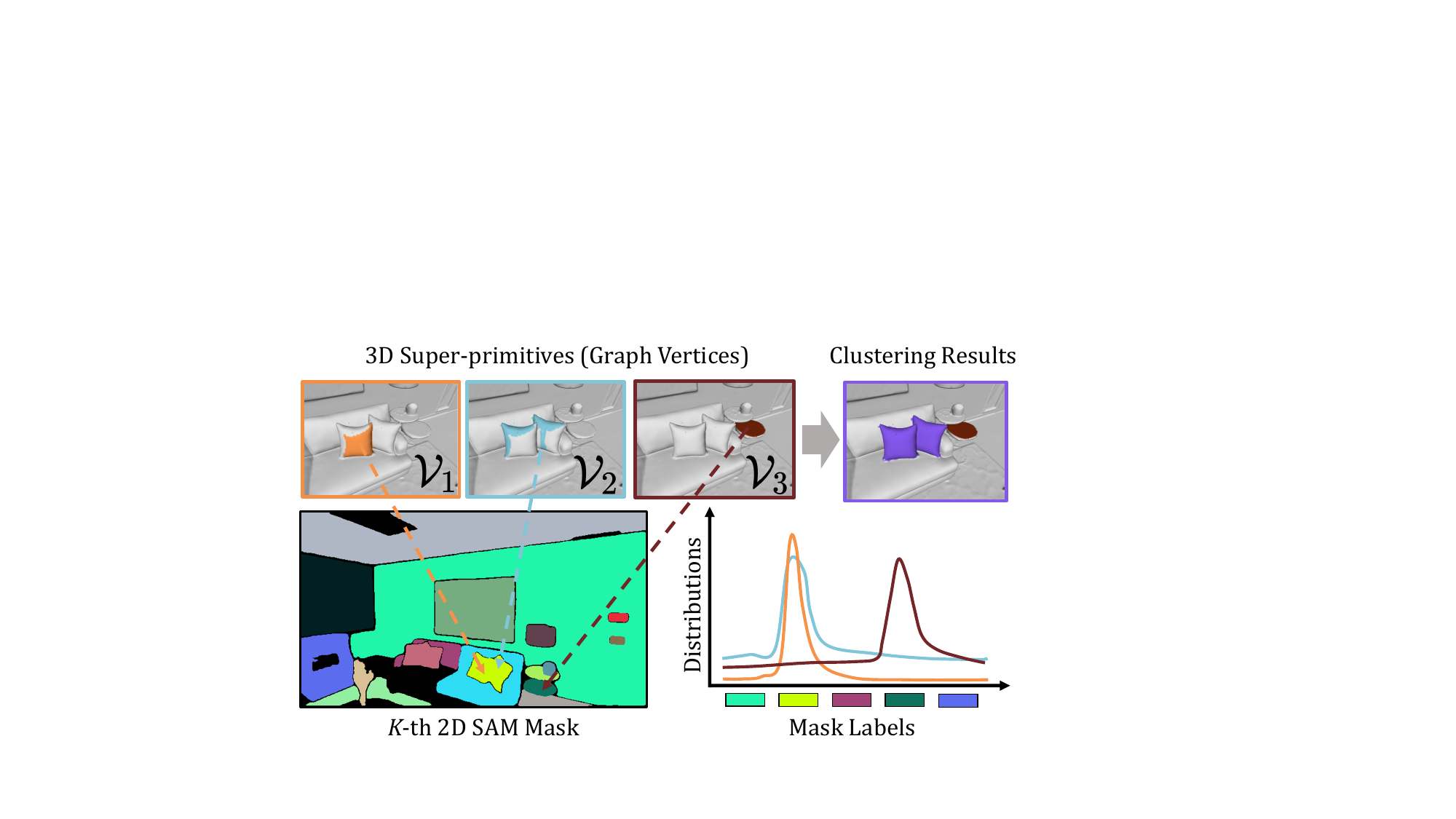} 
\caption{By projecting primitives inside different graph vertices into $k$-th 2D SAM mask, $\mathcal{V}_1$ and $\mathcal{V}_2$ have similar mask label distributions, while $\mathcal{V}_3$ has different mask label distributions. So, we can cluster $\mathcal{V}_1$ and $\mathcal{V}_2$ into the same category based on the distance between the distributions.}
\label{fig:clustering}
\end{figure}
\myvspace\noindent\textbf{Edge Affinity Computation.}
After obtaining graph vertices $\{\mathcal{V}_i\}$, we build edge and compute affinity based on the spatial adjacency relationships between these super-primitives.
Inspired by recent works~\cite{wu2024opengaussian,takmaz2023openmask3d,yin2024sai3d,yan2024maskclustering,gaussian_grouping}
that use powerful segmentation models~\cite{sam,cropformer} to generate multi-view 2D instance masks as the guidance of instance-level feature learning and clustering. 
Similarly, we also use the 2D multi-view instance masks generated by SAM~\cite{sam} to aid the edge affinity computation.

To build the affinity between two super-primitives $\mathcal{V}_i$ and $\mathcal{V}_j$, we first project them into $k$-th image mask according to the camera intrinsic.
The 3D Gaussian primitives inside $\mathcal{V}_i$ will fall in $k$-th instance mask, and each primitive should have a 2D mask label in the $k$-th mask.
So, we can obtain its 2D mask label distribution $q_i$ based on the segmentation results of SAM~\cite{sam}.
As shown in~\cref{fig:clustering}, two super-primitives that belong to the same 2D instance should have similar 2D mask label distributions.
So, we can take the 2D mask label distribution as our 3D super-primitives consistency criterion.
To measure the difference between the 2D mask label distribution of two 3D super-primitives, we use the Jensen-Shannon divergence (JSD):
\begin{equation}
    \mathcal{E}_{ij}^{k} = \frac{1}{2} \sum_{z} [q_i(z)\log(\frac{q_i(z)}{y(z)}) + q_j(z)\log(\frac{q_j(z)}{y(z)})],
    \label{eq:single_view_jsd}
\end{equation}
where $y=(q_i+q_j)/2$ is the average of two mask label distributions.

Evidently, affinity from a single view may be noisy and inaccurate. 
We should consider the multi-view consistency when performing 3D consistent panoptic segmentation.
So, we aggregate multi-view affinity to achieve cross-view consistency based on the proportion of primitives that can be observed in different views.
The final multi-view affinity is defined in the following equation:
\begin{equation}
    \mathcal{E}_{ij} = \frac{1}{k}\sum_{k=1} (\frac{\texttt{Vis}(\mathcal{V}_i)}{|\mathcal{V}_i|} \cdot \frac{\texttt{Vis}(\mathcal{V}_j)}{|\mathcal{V}_j|} \cdot \mathcal{E}_{ij}^{k})
    \label{eq:multi_view_jsd}
\end{equation}
where $\texttt{Vis}(\cdot)$ function indicates the number of visible primitives in the current viewpoint and $|\mathcal{V}_i|$ denotes the number of primitives insides $\mathcal{V}_i$.

When we obtain the multi-view consistency criterion of two super-primitives, we can merge them into the same instance group based on the following progressive clustering process.

\myvspace\noindent\textbf{Progressive Graph Clustering.}
According to the graph $\mathcal{G} = (\{\mathcal{V}_i\}, \{\mathcal{E}_{ij}\})$ built in the previous section, we perform graph clustering that merges $\{\mathcal{
V}_i\}$ with large affinity scores into the same instance.
To obtain the global consistent clustering results, we adopt a progressive local-to-global way to merge super-primitives with spatial neighbor connections.
Specifically, we first cluster the local super-primitives with high-affinity scores and merge them into large super-primitives.
During each iteration, we update the graph vertexes and edge affinity for the next iteration due to the changes in the scene graph structure.
We update the affinity threshold during progressive clustering, which is linearly reduced from 0.9 to 0.6 with 4 iterations.

\subsection{Open Vocabulary Panoptic Segmentation}
\label{subsec:openset_query}
When we finish progressive graph clustering, we can obtain a set of non-overlapping clustering groups where each entry represents a 3D class-agnostic instance.
Besides, with the 3D feature decoder optimized in~\cref{subsec:feature_learning}, we can obtain the 3D language feature for each primitive.
For 3D open vocabulary panoptic segmentation, we assume that the Gaussian primitives inside the same super-primitive should belong to the same semantic category. 
Therefore, we use a prediction voting method to calculate the semantic category of the super-primitive, which can get more complete instance-level semantic segmentation results.

\section{Experiments}
\label{sec:exps}
In this section, we first describe our experimental setting and then present quantitative and qualitative results of our approach and state-of-the-art baselines on two commonly used datasets. 
Additionally, we perform a detailed ablation study to justify our design choices.

\subsection{Experimental Settings}
\myvspace\noindent\textbf{Datasets.}
Following the previous works~\cite{wu2024opengaussian,Peng2023OpenScene}, we evaluate our method on two widely used indoor scene datasets, Replica~\cite{julian:2019:replica} and ScanNetV2~\cite{dai:2017:scannet} for both quantitative and qualitative evaluations.
The Replica dataset contains high-quality indoor scenes with carefully annotated ground-truth semantic and instance labels.
For a fair comparison, we take the commonly-used 8 scenes $\{\texttt{room0-2,office0-4}\}$ for evaluation.
The ScanNetV2~\cite{dai:2017:scannet} dataset consists various of challenging indoor scenes with different numbers of RGB-D frames for each sequence, as well as reconstructed point clouds and GT 3D point-level semantic labels.
Following~\cite{wu2024opengaussian}, we use the same 10 selected sequences and settings for the evaluation of scene understanding.

\myvspace\noindent\textbf{Evaluation Metrics.}
For the evaluation of open vocabulary 3D panoptic segmentation, we evaluate the performance of our method on four widely-used metrics: point cloud mean Intersection over Union (mIoU), mean Accuracy (mAcc.), and 3D Panoptic Reconstruction Quality (PRQ)~\cite{dahnert2021single_panoptic} which is modified from the common 2D panoptic segmentation quality~\cite{kirillov2019panoptic}.
In our experiments, we use the \textit{thing}-level metric, PRQ (T) and \textit{stuff}-level metric, PRQ (S) for evaluation.

\myvspace\noindent\textbf{Implementation Details.}
Following~\cite{Peng2023OpenScene,wu2024opengaussian}, we use LSeg~\cite{Lseg} as our pixel-aligned visual-language feature extractors for ScanNetV2 and Replica datasets.
To extract the language feature of the text query, we use the OpenCLIP~\cite{clip} ViT-B/16 model.
For the 2D masks, we use the ViT-H SAM~\cite{sam} model to segment images. 
More details are provided in our supplementary material.

\myvspace\noindent\textbf{Baselines.}
We compare our approach with recent 3DGS-based approaches, LangSplat~\cite{qin2024langsplat}, OpenGaussians~\cite{wu2024opengaussian} and point-cloud based method, OpenScene~\cite{Peng2023OpenScene}.
Due to the performance of LangSplat~\cite{qin2024langsplat} reported in~\cite{wu2024opengaussian} is too worse, we modify it for better performance, which is indicated by $*$. And $\dag$ indicates that no Gaussian filtering is used in~\cite{wu2024opengaussian} for the evaluation of panoptic segmentation.
Besides, due to these methods can only extract the point-level language features, we use the fully supervised 3D instance segmentation approach SoftGroup~\cite{vu2022softgroup} (trained on ScanNetV2~\cite{dai:2017:scannet}) to provide instance mask proposals for the comparison of panoptic segmentation.


\begin{table}[t]
\centering
\small
\setlength{\tabcolsep}{0.6em}
\vspace{-2mm}
\begin{tabular}{lcccc}
\toprule
    Method & mIoU & mAcc. & PRQ (T) & PRQ (S)\\
    \midrule
    \multicolumn{4}{l}{\textit{Open-vocab. semantic + Sup. mask~\cite{vu2022softgroup}}} \\
    LangSplat~\cite{qin2024langsplat} & 3.78 & 9.11 & --- & --- \\
    LangSplat$^{*}$~\cite{qin2024langsplat} & 29.47 & 45.29 & 22.57 & 28.44 \\
    OpenGaussian~\cite{wu2024opengaussian} & 24.73 & 41.54 & --- & --- \\
    OpenGaussian$^{\dag}$~\cite{wu2024opengaussian} & 24.89 & 37.35 & 22.87 & 19.71 \\
    OpenScene($\textit{Dis.}$)~\cite{Peng2023OpenScene} & 46.91 & 68.50 & \underline{43.77} & \underline{40.69} \\
    OpenScene($\textit{Ens.}$)~\cite{Peng2023OpenScene} & \underline{47.63} & \underline{69.74} & 43.53 & 40.43 \\
    \noalign{\vskip 1pt} \hdashline \noalign{\vskip 1pt}
    Ours & \textbf{50.72} & \textbf{70.20} & 33.84 & 36.22 \\
    Ours \textit{+ Sup. mask~\cite{vu2022softgroup}} & \textbf{50.72} & \textbf{70.20} & \textbf{49.26} & \textbf{48.24} \\
\bottomrule
\end{tabular}
\caption{3D semantic and panoptic segmentation results on ScanNetV2~\cite{dai:2017:scannet}. The results of~\cite{qin2024langsplat} and~\cite{wu2024opengaussian} are taken from~\cite{wu2024opengaussian} and OpenScene~\cite{Peng2023OpenScene} are obtained from their pre-trained model. $*$ indicates our better implementation. $\dag$ indicates no Gaussian filter is used for the evaluation of panoptic segmentation.}
\label{tab:scannet_results}
\end{table}


\begin{table}[t]
\centering
\small
\setlength{\tabcolsep}{0.5em}
\begin{tabular}{lcccc}
\toprule
    Method & mIoU & mAcc. & PRQ (T) & PRQ (S)\\
    \midrule
    \multicolumn{4}{l}{\textit{Open-vocab. semantic + Sup. mask~\cite{vu2022softgroup}}}  \\
    LangSplat$^{*}$~\cite{qin2024langsplat} & 4.82 & 10.03 & 8.29 & 1.28 \\
    OpenScene($\textit{Dis.}$)~\cite{Peng2023OpenScene} & 44.32 & 56.14 & 31.43 & 10.95 \\
    OpenScene($\textit{Ens.}$)~\cite{Peng2023OpenScene} & \underline{49.03} & \underline{62.89} & 33.04 & \underline{11.84} \\
    \noalign{\vskip 1pt} \hdashline \noalign{\vskip 1pt}
    Ours  & \textbf{54.98} & \textbf{67.35} & \textbf{43.04} & \textbf{30.60} \\
    Ours + \textit{Sup. mask~\cite{vu2022softgroup}} & \textbf{54.98} & \textbf{67.35} & \underline{40.80} & 11.31 \\
\bottomrule
\end{tabular}
\caption{3D semantic and panoptic segmentation results on Replica~\cite{julian:2019:replica}. The results of OpenScene~\cite{Peng2023OpenScene} are obtained from their pre-trained model. $*$ indicates our better implementation of LangSplat~\cite{qin2024langsplat}.}
\label{tab:replica_results}
\end{table}
\begin{figure*}[ht!]
\centering
 \includegraphics[width=\linewidth]{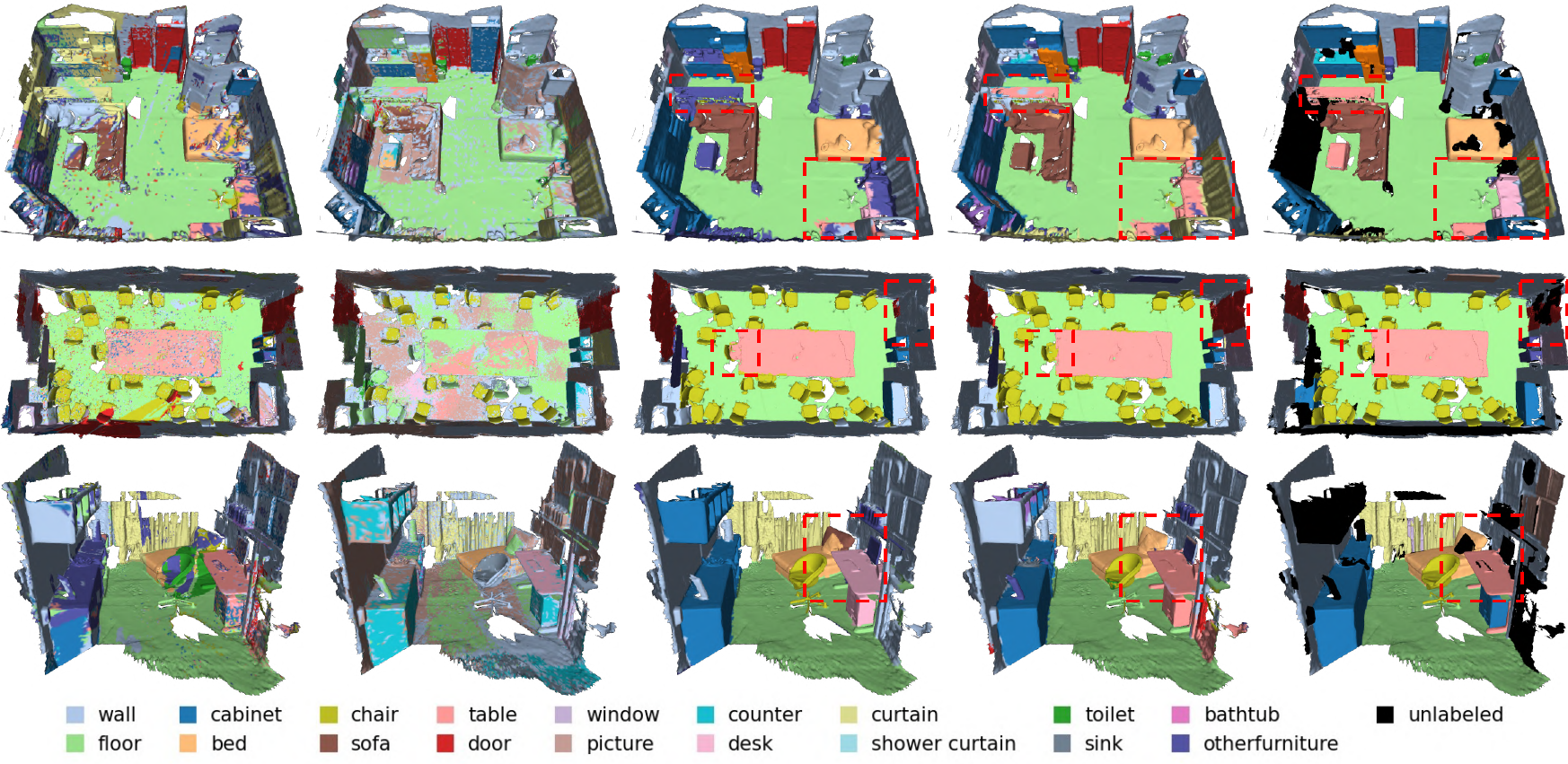} 
 \vspace{1em}
 \begin{minipage}{0.19\linewidth}
 \centering
 \small
 LangSplat~\cite{qin2024langsplat}
 \end{minipage}
 \hfill
  \begin{minipage}{0.2\linewidth}
 \centering
 \small
 OpenGaussian~\cite{wu2024opengaussian}
 \end{minipage}
 \hfill
  \begin{minipage}{0.2\linewidth}
 \centering
 \small
 OpenScene~\cite{Peng2023OpenScene}
 \end{minipage}
 \hfill
  \begin{minipage}{0.2\linewidth}
 \centering
 \small
 \textbf{Ours}
 \end{minipage}
 \hfill
  \begin{minipage}{0.19\linewidth}
 \centering
 \small
 GT Segmentation
 \end{minipage}
\vspace{-5mm}
\caption{Qualitative 3D semantic segmentation comparison of ScanNetV2~\cite{dai:2017:scannet}. Our approach outperforms recent 3DGS-based approaches, LangSplat~\cite{qin2024langsplat} and OpenGaussian~\cite{wu2024opengaussian}, by a large margin. Compared with OpenScene~\cite{Peng2023OpenScene}, we can achieve better segmentation results on thing-level objects.}
\label{fig:scannet_semantic_mesh}
\end{figure*}
\begin{figure}[h]
\centering
\includegraphics[width=\linewidth]{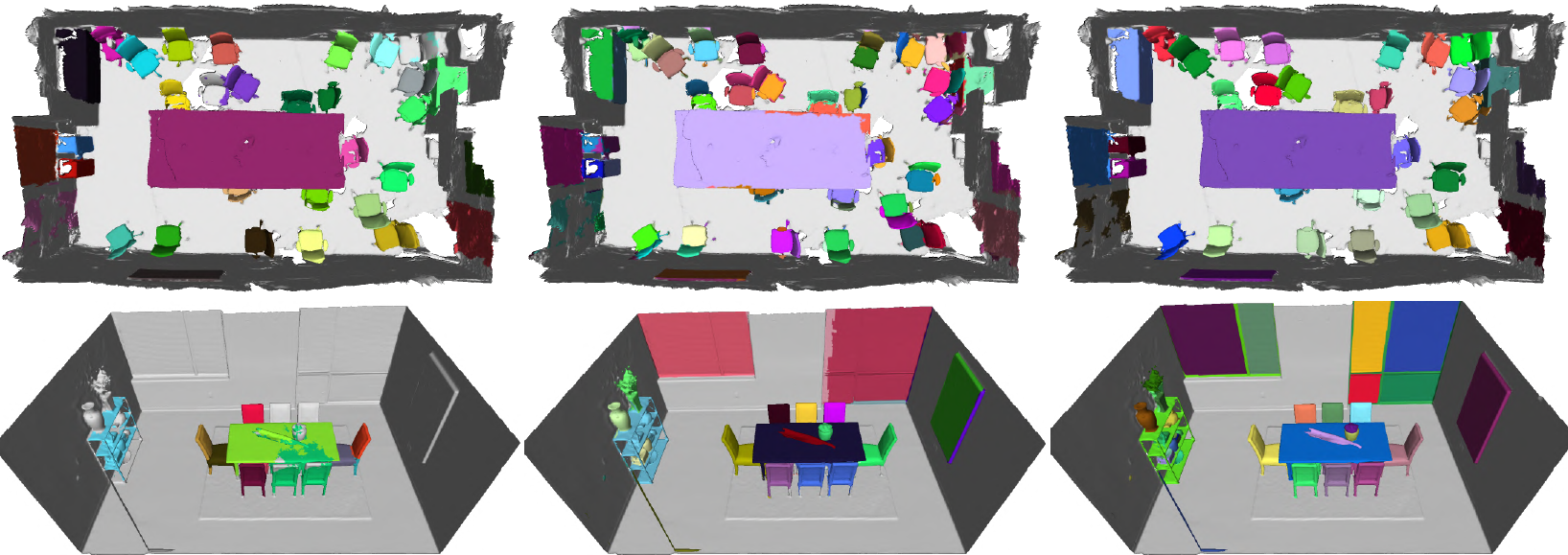} 
 \begin{minipage}{0.32\linewidth}
 \centering
 \small
 SoftGroup~\cite{vu2022softgroup}
 \end{minipage}
 \hfill
  \begin{minipage}{0.32\linewidth}
 \centering
 \small
 \textbf{Ours}
 \end{minipage}
 \hfill
 \begin{minipage}{0.32\linewidth}
 \centering
 \small
 GT Panoptic
 \end{minipage}
\caption{Qualitative 3D panoptic segmentation comparison. We show two reconstructed panoptic maps selected from ScanNetV2~\cite{dai:2017:scannet} and Replica~\cite{julian:2019:replica} datasets.}
\label{fig:exp_panoptic_mesh}
\end{figure}
\begin{figure}[h]
\centering
\includegraphics[width=\linewidth]{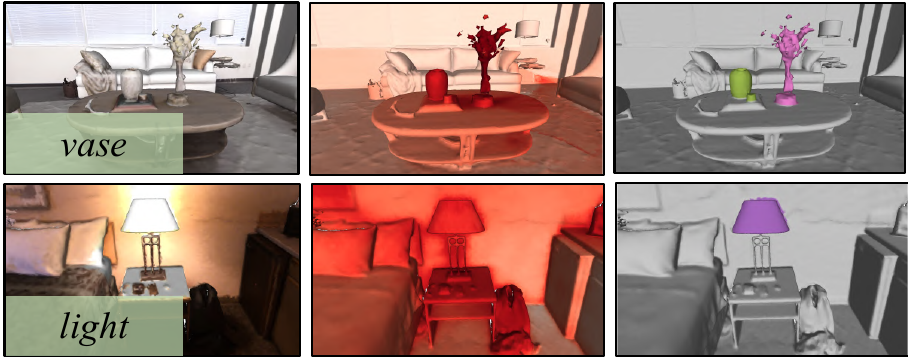} 
 \begin{minipage}{0.32\linewidth}
 \centering
 \small
 RGB \& Query
 \end{minipage}
 \hfill
  \begin{minipage}{0.32\linewidth}
 \centering
 \small
 OpenScene~\cite{Peng2023OpenScene}
 \end{minipage}
 \hfill
  \begin{minipage}{0.32\linewidth}
 \centering
 \small
 \textbf{Ours}
 \end{minipage}
\caption{Qualitative results of open vocabulary query. The query text is in the lower left of the RGB image. For OpenScene, the redder the color, the higher the similarity. We use different colors to distinguish different instances found in the query.}
\label{fig:exp_open_query}
\end{figure}

\subsection{Main Experiments}
We evaluate the 3D panoptic segmentation metrics of our approach and baseline on two commonly used ScanNetV2~\cite{dai:2017:scannet} and Replica~\cite{julian:2019:replica} datasets.
Due to the open-source codes of OpenGaussian~\cite{wu2024opengaussian} is not complete, we can't obtain its 3D scene understanding performance on the Replica dataset.
For OpenScene~\cite{Peng2023OpenScene}
, we use its 3D Distill (\textit{Dis.}) and 2D/3D Ensemble (\textit{Ens.}) variants for comparison. 
The best results are shown in \textbf{bolded}, and the second-best results are represented in \underline{underlined}.

\myvspace\noindent\textbf{3D Semantic Segmentation.}
The averaged quantitative semantic segmentation results are shown in~\cref{tab:scannet_results} and~\cref{tab:replica_results}, respectively.
According to the 3D semantic segmentation results in the table, our approach achieves the best results on the mIoU and mAcc metrics of two datasets.
Compared with 3DGS-based approaches~\cite{qin2024langsplat,wu2024opengaussian}, our method can achieve significant improvements.
Benefiting from learning language features in the 3D space, we can learn consistent features for each Gaussian primitive, and avoid the bias introduced by alpha-blending.
Previous 3DGS-based approaches learn individual and discrete features for each Gaussian primitive which can lead to spatial noise and destroy the smoothness of the semantic features.
Our approach inherently learns language features from a 3D latent pyramid tri-plane.
Also, OpenScene~\cite{Peng2023OpenScene} which utilizes the 3D and 2D information to perform scene understanding has achieved better performance than current 3DGS-based methods.
Our method still performs better than OpenScene on the open vocabulary 3D semantic segmentation task.
The qualitative semantic segmentation results with open vocabulary query of ScanNetV2~\cite{dai:2017:scannet} are shown in~\cref{fig:scannet_semantic_mesh}. Our method can achieve consistent segmentation results, while previous 3DGS-based methods often lead to noisy segmentation results, as shown in the first two columns of~\cref{fig:scannet_semantic_mesh}. 
In addition, compared with OpenScene, we can achieve better segmentation results on the long-tail categories.

\myvspace\noindent\textbf{3D Panoptic Segmentation.}
The averaged quantitative results are also shown in~\cref{tab:scannet_results} and~\cref{tab:replica_results}, respectively.
For the 3D panoptic segmentation performance, due to the compared baselines only output point-level segmentation results, we use the fully supervised 3D instance segmentation approach, SoftGroup~\cite{vu2022softgroup} (trained on ScanNetV2), to generate 3D instance proposals for them.
Since SoftGroup~\cite{vu2022softgroup} has a better instance segmentation performance on ScanNetV2, our PRQ (T) and PRQ (S) are slightly worse than the combination of OpenScene+SoftGroup. 
However, when we use the 3D instance maks generated by SoftGroup, we achieve the best results on 3D panoptic reconstruction quality.
For the Replica dataset, the segmentation results of SoftGroup are worse than our approach. 
Based on our learned language features and clustering-based segmentation results, we can achieve the best results in terms of PRQ (T) and PRQ (S). 
Besides, we also show the qualitative panoptic segmentation results of SoftGroup~\cite{vu2022softgroup} and ours in~\cref{fig:exp_panoptic_mesh}.
As can be seen from the figure, we can generalize better than SoftGroup~\cite{vu2022softgroup} for the instance segmentation results on the Replica~\cite{julian:2019:replica} dataset.

\myvspace\noindent\textbf{Open Vocabulary Query.}
We show the qualitative open vocabulary query results of two selected scenes in~\cref{fig:exp_open_query}.
Previous methods~\cite{qin2024langsplat,wu2024opengaussian,Peng2023OpenScene,shi2024_gs_language_embed} calculated the similarity between scene features and text queries, and they can not distinguish different objects with the same semantics. 
However, our approach can obtain instance-level information of different objects with the same semantics through our clustering (shown in the first row of~\cref{fig:exp_open_query}).

\subsection{Ablation Studies and Analysis}
We conduct ablation studies to analyze the effectiveness of 3D language feature learning and graph clustering based segmentation modules.
When we analyze one module, the other one is fixed and under the default setting.

\begin{figure}[h]
\centering
\includegraphics[width=0.49\linewidth]{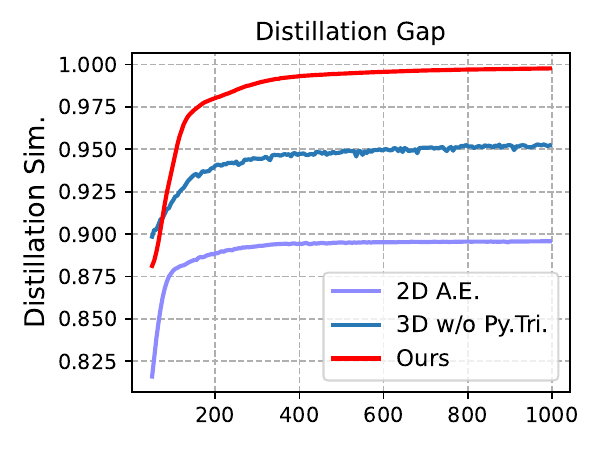}
\includegraphics[width=0.49\linewidth]{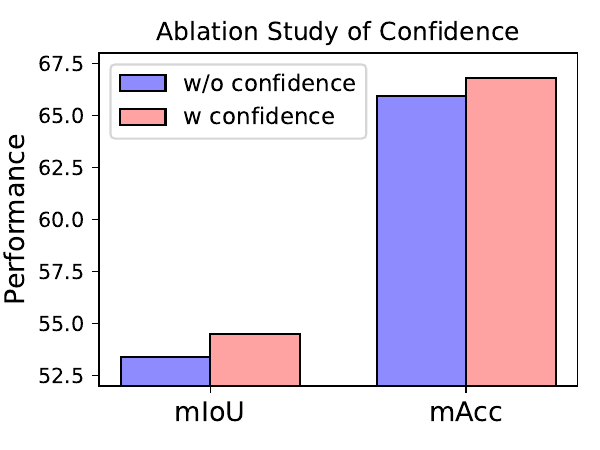}
\caption{Ablation studies of our 3D language learning module. The results are evaluated on Replica~\cite{julian:2019:replica} dataset.}
\label{fig:ablation_cosine_confidence}
\end{figure}

\begin{table}[t]
\centering
\small
\setlength{\tabcolsep}{0.5em}
\begin{tabular}{cccccc}
\toprule
\makecell{\textit{3D Dec.}} & \makecell{\textit{Py. Tri.}} & mIoU & mAcc. & PRQ (T) & PRQ (S) \\
\midrule
\ding{55} & \ding{55} & 24.51 & 40.48 & 20.57 & 23.07 \\
\ding{51} & \ding{55} & 36.39 & 53.17 & 23.68 & 33.59 \\
\ding{51} & \ding{51} & \textbf{50.72} & \textbf{70.20} & \textbf{33.84} & \textbf{36.22} \\
\bottomrule
\end{tabular}
\caption{Ablation studies of our 3D language feature learning module. The results are evaluated on the ScanNetV2~\cite{dai:2017:scannet} dataset.}
\label{tab:ablation_feature_learning}
\end{table}
\myvspace\noindent\textbf{Effects of the designs for our 3D language feature learning module.} In~\cref{tab:ablation_feature_learning}, we show the quantitative analysis of our feature learning module.
\textit{3D} \textit{Dec.} and \textit{Py.} \textit{Tir.} indicate using the 3D decoder to regress language features from the projected multi-view primitive-level feature and using our latent pyramid tri-planes, respectively.
w/o \textit{3D Dec.} and \textit{Py. Tri.} means that we use a 2D autoencoder similar to LangSplat~\cite{qin2024langsplat} and replace triplane with positional encoding for feature learning.
As shown in the table, compared with directly performing distillation with 2D auto encode-decoder (without \textit{3D Dec.} and \textit{Py. Tri.}), using 3D feature distillation and latent parametric encoding both can lead to better performance and reduce the noise introduced by the discrete Gaussian language feature.
Additionally, on the left of~\cref{fig:ablation_cosine_confidence}, we also show the language feature learning gap of using 2D and 3D distillation way. 
From the figure, we can know that for large scenes, using the 2D distillation, the similarity of the learned language feature can only reach 0.9. 
But using the coordinate-based 3D distillation approach and our latent pyramid tri-plane, the feature learning performance can reach 0.95 and close to 1, respectively.
Besides, the performance of using the confidence during distillation is shown in the right part of~\cref{fig:ablation_cosine_confidence}. 
The results also validate the effectiveness of our confidence based feature learning, which can efficiently reduce the impact of features from unobserved and unreliable areas.

\begin{table}[t]
\centering
\small
\setlength{\tabcolsep}{0.5em}
\begin{tabular}{ccccccc}
\toprule
\makecell{\textit{JSD.}} & \makecell{\textit{Lang.}} & \makecell{\textit{Vot.}} & mIoU & mAcc. & PRQ (T) & PRQ (S) \\
\midrule
\ding{55} & \ding{55} & \ding{55} & 50.21 & 61.69 & 20.28 & 17.42 \\
\ding{51} & \ding{55} & \ding{55} & 50.21 & 61.69 & 33.22 & 25.21 \\
\ding{51} & \ding{51} & \ding{55} & 50.21 & 61.69 & 35.28 & 27.41 \\
\ding{51} & \ding{55} & \ding{51} & 52.69 & 63.25 & 40.10 & \textbf{31.57} \\
\ding{51} & \ding{51} & \ding{51} & \textbf{54.98} & \textbf{67.35} & \textbf{43.04} & 30.60 \\
\bottomrule
\end{tabular}
\caption{Ablation studies of our graph clustering based segmentation module. The results are evaluated on the Replica~\cite{julian:2019:replica} dataset.}
\label{tab:ablation_sem_seg}
\end{table}

\begin{figure}[h]
\centering
\includegraphics[width=\linewidth]{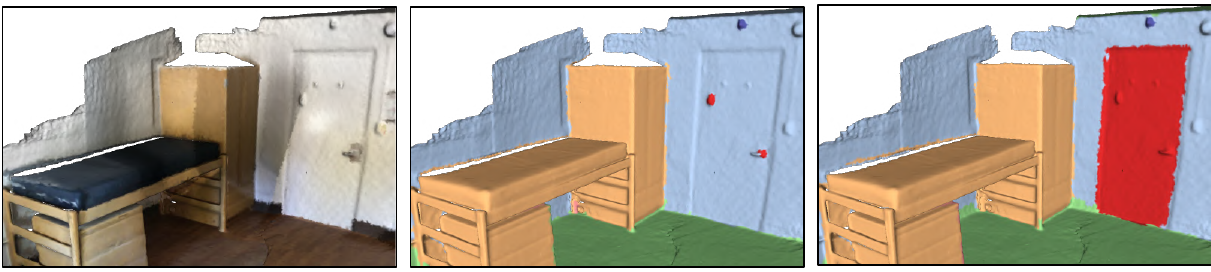} 
 \begin{minipage}{0.32\linewidth}
 \centering
 \small
 RGB
 \end{minipage}
 \hfill
  \begin{minipage}{0.32\linewidth}
 \centering
 \small
 w/o \textit{Lang.}
 \end{minipage}
 \hfill
 \begin{minipage}{0.32\linewidth}
 \centering
 \small
 w \textit{Lang.}
 \end{minipage}
\vspace{-3mm}
\caption{Qualitative comparison of using language-guided graph cut for graph vertex construction. 
Using our learned language feature can distinguish the door (shown in \red{red} color).}
\label{fig:ablation_lang_graph_cut}
\end{figure}
\myvspace\noindent\textbf{Effects of the designs for our graph clustering based segmentation module.} 
In~\cref{tab:ablation_sem_seg}, we show the quantitative analysis of our graph clustering based segmentation module.
\textit{JSD.}, \textit{Lang.}, and \textit{Vot.} indicate using multi-view JSD of mask label distributions to construct graph edge affinity, using language-guided graph cuts to construct graph vertices, and using predictions voting (\cref{subsec:openset_query}) for semantic segmentation, respectively.
Besides, w/o \textit{JSD.} means that we use the similarity of learned language features between super-primitives to construct edge affinity.
As can be seen from the table, compared with using language features (w/o \textit{JSD.}) to build graph edge affinity, using our multi-view affinity from mask label distributions will significantly improve our 3D panoptic reconstruction quality (PRQ (T): 33.22 \textit{v.s} 20.28 and PRQ (S): 25.21 \textit{v.s} 17.42).
Using our language-guided graph cuts can avoid merging different semantic objects with similar structures into the same vertex/instance, such as wall and door, as shown in~\cref{fig:ablation_lang_graph_cut}. 
It also can further improve the panoptic segmentation performance of 3D instances.
In addition, after obtaining accurate clustering results, when performing semantic segmentation, we vote the prediction results inside the same super-primitive to obtain consistent prediction results for objects.
The results validate \textit{Vot.} is also effective.
It can ensure that the primitives inside a 3D instance object have consistent semantic prediction results, which can increase instance-level IoU metrics for the predicted results and ground truth in 3D panoptic segmentation.

\section{Conclusion}
\label{sec:conclusion}
In this paper, we propose {\ours}, an effective 3DGS-based approach for 3D open vocabulary panoptic scene understanding which addresses the challenge of accurate 3D language feature learning and consistent instance-level open vocabulary segmentation.
For the semantic information, we regress the 3D language features from a latent continuous parametric feature space learned by the latent pyramid tri-planes and 3D feature decoder.
For the panoptic information, we adopt the language-guided graph cuts and progressive clustering strategy to construct geometrically and semantically consistent super-primitives and obtain the 3D panoptic information.
Extensive experiments on commonly used datasets demonstrate that {\ours} outperforms existing state-of-the-art methods for 3D open vocabulary panoptic scene understanding.

\noindent\textbf{Acknowledgment:} This work was partially supported by NSF of China (No. 62425209).


{
    \small
    \bibliographystyle{ieeenat_fullname}

\begin{thebibliography}{62}
\providecommand{\natexlab}[1]{#1}
\providecommand{\url}[1]{\texttt{#1}}
\expandafter\ifx\csname urlstyle\endcsname\relax
  \providecommand{\doi}[1]{doi: #1}\else
  \providecommand{\doi}{doi: \begingroup \urlstyle{rm}\Url}\fi

\bibitem[Bhalgat et~al.(2024)Bhalgat, Laina, Henriques, Zisserman, and Vedaldi]{bhalgat2024n2f2}
Yash Bhalgat, Iro Laina, Jo{\~{a}}o~F. Henriques, Andrew Zisserman, and Andrea Vedaldi.
\newblock {N2F2:} hierarchical scene understanding with nested neural feature fields.
\newblock In \emph{European Conference on Computer Vision}, pages 197--214, 2024.

\bibitem[Caron et~al.(2021)Caron, Touvron, Misra, J{\'e}gou, Mairal, Bojanowski, and Joulin]{dino}
Mathilde Caron, Hugo Touvron, Ishan Misra, Herv{\'e} J{\'e}gou, Julien Mairal, Piotr Bojanowski, and Armand Joulin.
\newblock Emerging properties in self-supervised vision transformers.
\newblock In \emph{IEEE/CVF International Conference on Computer Vision}, pages 9650--9660, 2021.

\bibitem[Chan et~al.(2022)Chan, Lin, Chan, Nagano, Pan, De~Mello, Gallo, Guibas, Tremblay, Khamis, et~al.]{chan2022eg3d}
Eric~R Chan, Connor~Z Lin, Matthew~A Chan, Koki Nagano, Boxiao Pan, Shalini De~Mello, Orazio Gallo, Leonidas~J Guibas, Jonathan Tremblay, Sameh Khamis, et~al.
\newblock Efficient geometry-aware 3d generative adversarial networks.
\newblock In \emph{IEEE/CVF Conference on Computer Vision and Pattern Recognition}, pages 16123--16133, 2022.

\bibitem[Chen et~al.(2024)Chen, Li, Ye, Wang, Xie, Zhai, Wang, Liu, Bao, and Zhang]{chen2024pgsr}
Danpeng Chen, Hai Li, Weicai Ye, Yifan Wang, Weijian Xie, Shangjin Zhai, Nan Wang, Haomin Liu, Hujun Bao, and Guofeng Zhang.
\newblock {PGSR}: Planar-based gaussian splatting for efficient and high-fidelity surface reconstruction.
\newblock \emph{IEEE Transactions on Visualization and Computer Graphics}, pages 1--12, 2024.

\bibitem[Cheng et~al.(2020)Cheng, Collins, Zhu, Liu, Huang, Adam, and Chen]{cheng2020panoptic}
Bowen Cheng, Maxwell~D Collins, Yukun Zhu, Ting Liu, Thomas~S Huang, Hartwig Adam, and Liang-Chieh Chen.
\newblock {Panoptic-DeepLab}: A simple, strong, and fast baseline for bottom-up panoptic segmentation.
\newblock In \emph{IEEE/CVF Conference on Computer Vision and Pattern Recognition}, pages 12475--12485, 2020.

\bibitem[Cheng et~al.(2022)Cheng, Misra, Schwing, Kirillov, and Girdhar]{cheng2021mask2former}
Bowen Cheng, Ishan Misra, Alexander~G. Schwing, Alexander Kirillov, and Rohit Girdhar.
\newblock Masked-attention mask transformer for universal image segmentation.
\newblock In \emph{IEEE/CVF Conference on Computer Vision and Pattern Recognition}, 2022.

\bibitem[Dahnert et~al.(2021)Dahnert, Hou, Nie{\ss}ner, and Dai]{dahnert2021single_panoptic}
Manuel Dahnert, Ji Hou, Matthias Nie{\ss}ner, and Angela Dai.
\newblock Panoptic 3d scene reconstruction from a single rgb image.
\newblock \emph{Advances in Neural Information Processing Systems}, 34:\penalty0 8282--8293, 2021.

\bibitem[Dai et~al.(2017)Dai, Chang, Savva, Halber, Funkhouser, and Nießner]{dai:2017:scannet}
Angela Dai, Angel~X. Chang, Manolis Savva, Maciej Halber, Thomas Funkhouser, and Matthias Nießner.
\newblock Scan{N}et: Richly-annotated {3D} reconstructions of indoor scenes.
\newblock In \emph{IEEE/CVF Conference on Computer Vision and Pattern Recognition}, pages 2432--2443, 2017.

\bibitem[Felzenszwalb and Huttenlocher(2004)]{felzenszwalb2004efficient_graph_cut}
Pedro~F Felzenszwalb and Daniel~P Huttenlocher.
\newblock Efficient graph-based image segmentation.
\newblock \emph{International journal of computer vision}, 59:\penalty0 167--181, 2004.

\bibitem[Fu et~al.(2022)Fu, Zhang, Chen, Lu, Zhu, Zhou, Geiger, and Liao]{fu2022panopticnerf}
Xiao Fu, Shangzhan Zhang, Tianrun Chen, Yichong Lu, Lanyun Zhu, Xiaowei Zhou, Andreas Geiger, and Yiyi Liao.
\newblock Panoptic {NeRF}: {3D}-to-{2D} label transfer for panoptic urban scene segmentation.
\newblock In \emph{International Conference on 3D Vision}, pages 1--11, 2022.

\bibitem[Gasperini et~al.(2021)Gasperini, Mahani, Marcos-Ramiro, Navab, and Tombari]{gasperini2021panoster}
Stefano Gasperini, Mohammad-Ali~Nikouei Mahani, Alvaro Marcos-Ramiro, Nassir Navab, and Federico Tombari.
\newblock Panoster: End-to-end panoptic segmentation of lidar point clouds.
\newblock \emph{IEEE Robotics and Automation Letters}, 6\penalty0 (2):\penalty0 3216--3223, 2021.

\bibitem[Guo et~al.(2024{\natexlab{a}})Guo, Zhu, Peng, Wang, Shen, Hu, and Zhou]{guo2022sam_graph}
Haoyu Guo, He Zhu, Sida Peng, Yuang Wang, Yujun Shen, Ruizhen Hu, and Xiaowei Zhou.
\newblock Sam-guided graph cut for 3d instance segmentation.
\newblock In \emph{European Conference on Computer Vision}, 2024{\natexlab{a}}.

\bibitem[Guo et~al.(2024{\natexlab{b}})Guo, Ma, Fan, Liu, and Li]{guo2024semantic_gaussian}
Jun Guo, Xiaojian Ma, Yue Fan, Huaping Liu, and Qing Li.
\newblock Semantic gaussians: Open-vocabulary scene understanding with 3d gaussian splatting, 2024{\natexlab{b}}.

\bibitem[Huang et~al.(2024)Huang, Yu, Chen, Geiger, and Gao]{Huang2DGS2024}
Binbin Huang, Zehao Yu, Anpei Chen, Andreas Geiger, and Shenghua Gao.
\newblock {2D} gaussian splatting for geometrically accurate radiance fields.
\newblock In \emph{SIGGRAPH}, 2024.

\bibitem[Huang et~al.(2023)Huang, Mees, Zeng, and Burgard]{huang23vlmaps}
Chenguang Huang, Oier Mees, Andy Zeng, and Wolfram Burgard.
\newblock Visual language maps for robot navigation.
\newblock In \emph{IEEE/CVF International Conference on Robotics and Automation}, 2023.

\bibitem[Kerbl et~al.(2023)Kerbl, Kopanas, Leimk{\"u}hler, and Drettakis]{kerbl3Dgaussians}
Bernhard Kerbl, Georgios Kopanas, Thomas Leimk{\"u}hler, and George Drettakis.
\newblock 3d gaussian splatting for real-time radiance field rendering.
\newblock \emph{ACM Transactions on Graphics}, 42\penalty0 (4), 2023.

\bibitem[Kerr et~al.(2023)Kerr, Kim, Goldberg, Kanazawa, and Tancik]{lerf}
Justin Kerr, Chung~Min Kim, Ken Goldberg, Angjoo Kanazawa, and Matthew Tancik.
\newblock {LERF:} language embedded radiance fields.
\newblock In \emph{{IEEE/CVF} International Conference on Computer Vision}, pages 19672--19682, 2023.

\bibitem[Kirillov et~al.(2019)Kirillov, He, Girshick, Rother, and Doll{\'a}r]{kirillov2019panoptic}
Alexander Kirillov, Kaiming He, Ross Girshick, Carsten Rother, and Piotr Doll{\'a}r.
\newblock Panoptic segmentation.
\newblock In \emph{IEEE/CVF Conference on Computer Vision and Pattern Recognition}, pages 9404--9413, 2019.

\bibitem[Kirillov et~al.(2023)Kirillov, Mintun, Ravi, Mao, Rolland, Gustafson, Xiao, Whitehead, Berg, Lo, et~al.]{sam}
Alexander Kirillov, Eric Mintun, Nikhila Ravi, Hanzi Mao, Chloe Rolland, Laura Gustafson, Tete Xiao, Spencer Whitehead, Alexander~C Berg, Wan-Yen Lo, et~al.
\newblock Segment anything.
\newblock In \emph{IEEE/CVF International Conference on Computer Vision}, pages 4015--4026, 2023.

\bibitem[Kundu et~al.(2022)Kundu, Genova, Yin, Fathi, Pantofaru, Guibas, Tagliasacchi, Dellaert, and Funkhouser]{kundu2022panopticneuralfiled}
Abhijit Kundu, Kyle Genova, Xiaoqi Yin, Alireza Fathi, Caroline Pantofaru, Leonidas~J Guibas, Andrea Tagliasacchi, Frank Dellaert, and Thomas Funkhouser.
\newblock Panoptic neural fields: A semantic object-aware neural scene representation.
\newblock In \emph{IEEE/CVF Conference on Computer Vision and Pattern Recognition}, pages 12871--12881, 2022.

\bibitem[Li et~al.(2022{\natexlab{a}})Li, Weinberger, Belongie, Koltun, and Ranftl]{Lseg}
Boyi Li, Kilian~Q Weinberger, Serge Belongie, Vladlen Koltun, and Rene Ranftl.
\newblock Language-driven semantic segmentation.
\newblock In \emph{International Conference on Learning Representations}, 2022{\natexlab{a}}.

\bibitem[Li et~al.(2022{\natexlab{b}})Li, Yang, Zhai, Liu, Bao, and Zhang]{Vox-Surf}
Hai Li, Xingrui Yang, Hongjia Zhai, Yuqian Liu, Hujun Bao, and Guofeng Zhang.
\newblock {Vox-Surf}: Voxel-based implicit surface representation.
\newblock \emph{IEEE Transactions on Visualization and Computer Graphics}, 30\penalty0 (3):\penalty0 1743--1755, 2022{\natexlab{b}}.

\bibitem[Li et~al.(2023)Li, Zhai, Yang, Wu, Zheng, Wang, Wu, Bao, and Zhang]{imtooth}
Hai Li, Hongjia Zhai, Xingrui Yang, Zhirong Wu, Yihao Zheng, Haofan Wang, Jianchao Wu, Hujun Bao, and Guofeng Zhang.
\newblock {ImTooth}: Neural implicit tooth for dental augmented reality.
\newblock \emph{{IEEE} Trans. Vis. Comput. Graph.}, 29\penalty0 (5):\penalty0 2837--2846, 2023.

\bibitem[Lu et~al.(2023)Lu, Kuen, Tiancheng, Jiuxiang, Weidong, Jiaya, Zhe, and Ming-Hsuan]{cropformer}
Qi Lu, Jason Kuen, Shen Tiancheng, Gu Jiuxiang, Guo Weidong, Jia Jiaya, Lin Zhe, and Yang Ming-Hsuan.
\newblock High-quality entity segmentation.
\newblock In \emph{IEEE/CVF International Conference on Computer Vision}, 2023.

\bibitem[Matsuki et~al.(2024)Matsuki, Murai, Kelly, and Davison]{monogs}
Hidenobu Matsuki, Riku Murai, Paul~HJ Kelly, and Andrew~J Davison.
\newblock Gaussian splatting slam.
\newblock In \emph{IEEE/CVF Conference on Computer Vision and Pattern Recognition}, pages 18039--18048, 2024.

\bibitem[Max(1995)]{max1995optical}
Nelson Max.
\newblock Optical models for direct volume rendering.
\newblock \emph{IEEE Transactions on Visualization and Computer Graphics}, 1\penalty0 (2):\penalty0 99--108, 1995.

\bibitem[Mildenhall et~al.(2020)Mildenhall, Srinivasan, Tancik, Barron, Ramamoorthi, and Ng]{mildenhall:2020:nerf}
Ben Mildenhall, Pratul~P. Srinivasan, Matthew Tancik, Jonathan~T. Barron, Ravi Ramamoorthi, and Ren Ng.
\newblock Nerf: Representing scenes as neural radiance fields for view synthesis.
\newblock In \emph{European Conference on Computer Vision}, 2020.

\bibitem[Ming et~al.(2025)Ming, Yang, Wang, Chen, Feng, Xing, and Zhang]{ming2024benchmarking}
Yuhang Ming, Xingrui Yang, Weihan Wang, Zheng Chen, Jinglun Feng, Yifan Xing, and Guofeng Zhang.
\newblock Benchmarking neural radiance fields for autonomous robots: An overview.
\newblock \emph{Engineering Applications of Artificial Intelligence}, 2025.

\bibitem[Narita et~al.(2019)Narita, Seno, Ishikawa, and Kaji]{narita2019panopticfusion}
Gaku Narita, Takashi Seno, Tomoya Ishikawa, and Yohsuke Kaji.
\newblock Panopticfusion: Online volumetric semantic mapping at the level of stuff and things.
\newblock In \emph{IEEE/RSJ International Conference on Intelligent Robots and Systems}, pages 4205--4212, 2019.

\bibitem[Peng et~al.(2023)Peng, Genova, Jiang, Tagliasacchi, Pollefeys, and Funkhouser]{Peng2023OpenScene}
Songyou Peng, Kyle Genova, Chiyu~"Max" Jiang, Andrea Tagliasacchi, Marc Pollefeys, and Thomas Funkhouser.
\newblock Openscene: 3d scene understanding with open vocabularies.
\newblock In \emph{IEEE/CVF Conference on Computer Vision and Pattern Recognition}, 2023.

\bibitem[Porzi et~al.(2019)Porzi, Bulo, Colovic, and Kontschieder]{porzi2019seamless}
Lorenzo Porzi, Samuel~Rota Bulo, Aleksander Colovic, and Peter Kontschieder.
\newblock Seamless scene segmentation.
\newblock In \emph{IEEE/CVF conference on computer vision and pattern recognition}, pages 8277--8286, 2019.

\bibitem[Qin et~al.(2024)Qin, Li, Zhou, Wang, and Pfister]{qin2024langsplat}
Minghan Qin, Wanhua Li, Jiawei Zhou, Haoqian Wang, and Hanspeter Pfister.
\newblock Langsplat: 3d language gaussian splatting.
\newblock In \emph{IEEE/CVF Conference on Computer Vision and Pattern Recognition}, pages 20051--20060, 2024.

\bibitem[Radford et~al.(2021)Radford, Kim, Hallacy, Ramesh, Goh, Agarwal, Sastry, Askell, Mishkin, Clark, et~al.]{clip}
Alec Radford, Jong~Wook Kim, Chris Hallacy, Aditya Ramesh, Gabriel Goh, Sandhini Agarwal, Girish Sastry, Amanda Askell, Pamela Mishkin, Jack Clark, et~al.
\newblock Learning transferable visual models from natural language supervision.
\newblock In \emph{International Conference on Machine Learning}, pages 8748--8763, 2021.

\bibitem[Shi et~al.(2024)Shi, Wang, Duan, and Guan]{shi2024_gs_language_embed}
Jin-Chuan Shi, Miao Wang, Hao-Bin Duan, and Shao-Hua Guan.
\newblock Language embedded 3d gaussians for open-vocabulary scene understanding.
\newblock In \emph{IEEE/CVF Conference on Computer Vision and Pattern Recognition}, pages 5333--5343, 2024.

\bibitem[Siddiqui et~al.(2023)Siddiqui, Porzi, Bul{\`o}, M{\"u}ller, Nie{\ss}ner, Dai, and Kontschieder]{siddiqui2023panoptic-lifting}
Yawar Siddiqui, Lorenzo Porzi, Samuel~Rota Bul{\`o}, Norman M{\"u}ller, Matthias Nie{\ss}ner, Angela Dai, and Peter Kontschieder.
\newblock Panoptic lifting for 3d scene understanding with neural fields.
\newblock In \emph{IEEE/CVF Conference on Computer Vision and Pattern Recognition}, pages 9043--9052, 2023.

\bibitem[Stearns et~al.(2024)Stearns, Harley, Uy, Dubost, Tombari, Wetzstein, and Guibas]{stearns2024dynamic_marbles}
Colton Stearns, Adam Harley, Mikaela Uy, Florian Dubost, Federico Tombari, Gordon Wetzstein, and Leonidas Guibas.
\newblock Dynamic gaussian marbles for novel view synthesis of casual monocular videos.
\newblock \emph{arXiv preprint arXiv:2406.18717}, 2024.

\bibitem[Straub et~al.(2019)Straub, Whelan, Ma, Chen, Wijmans, Green, Engel, Mur-Artal, Ren, Verma, Clarkson, Yan, Budge, Yan, Pan, Yon, Zou, Leon, Carter, Briales, Gillingham, Mueggler, Pesqueira, Savva, Batra, Strasdat, Nardi, Goesele, Lovegrove, and Newcombe]{julian:2019:replica}
Julian Straub, Thomas Whelan, Lingni Ma, Yufan Chen, Erik Wijmans, Simon Green, Jakob~J. Engel, Raul Mur-Artal, Carl Ren, Shobhit Verma, Anton Clarkson, Mingfei Yan, Brian Budge, Yajie Yan, Xiaqing Pan, June Yon, Yuyang Zou, Kimberly Leon, Nigel Carter, Jesus Briales, Tyler Gillingham, Elias Mueggler, Luis Pesqueira, Manolis Savva, Dhruv Batra, Hauke~M. Strasdat, Renzo~De Nardi, Michael Goesele, Steven Lovegrove, and Richard Newcombe.
\newblock The {R}eplica dataset: A digital replica of indoor spaces.
\newblock \emph{arXiv preprint arXiv:1906.05797}, 2019.

\bibitem[Takmaz et~al.(2023)Takmaz, Fedele, Sumner, Pollefeys, Tombari, and Engelmann]{takmaz2023openmask3d}
Ay{\c{c}}a Takmaz, Elisabetta Fedele, Robert~W. Sumner, Marc Pollefeys, Federico Tombari, and Francis Engelmann.
\newblock {OpenMask3D: Open-Vocabulary 3D Instance Segmentation}.
\newblock In \emph{Advances in Neural Information Processing Systems}, 2023.

\bibitem[Tang et~al.(2023)Tang, Ren, Zhou, Liu, and Zeng]{tang2023dreamgaussian}
Jiaxiang Tang, Jiawei Ren, Hang Zhou, Ziwei Liu, and Gang Zeng.
\newblock Dreamgaussian: Generative gaussian splatting for efficient 3d content creation.
\newblock \emph{arXiv preprint arXiv:2309.16653}, 2023.

\bibitem[Tschernezki et~al.(2022)Tschernezki, Laina, Larlus, and Vedaldi]{n3f}
Vadim Tschernezki, Iro Laina, Diane Larlus, and Andrea Vedaldi.
\newblock Neural feature fusion fields: 3d distillation of self-supervised 2d image representations.
\newblock In \emph{International Conference on 3D Vision}, pages 443--453, 2022.

\bibitem[Turkulainen et~al.(2024)Turkulainen, Ren, Melekhov, Seiskari, Rahtu, and Kannala]{turkulainen2024dn_splatter}
Matias Turkulainen, Xuqian Ren, Iaroslav Melekhov, Otto Seiskari, Esa Rahtu, and Juho Kannala.
\newblock Dn-splatter: Depth and normal priors for gaussian splatting and meshing.
\newblock \emph{arXiv preprint arXiv:2403.17822}, 2024.

\bibitem[Vu et~al.(2022)Vu, Kim, Luu, Nguyen, and Yoo]{vu2022softgroup}
Thang Vu, Kookhoi Kim, Tung~M Luu, Thanh Nguyen, and Chang~D Yoo.
\newblock Softgroup for 3d instance segmentation on point clouds.
\newblock In \emph{IEEE/CVF Conference on Computer Vision and Pattern Recognition}, pages 2708--2717, 2022.

\bibitem[Wang et~al.(2024{\natexlab{a}})Wang, Ye, Gao, Austin, Li, and Kanazawa]{shape-of-motion}
Qianqian Wang, Vickie Ye, Hang Gao, Jake Austin, Zhengqi Li, and Angjoo Kanazawa.
\newblock Shape of motion: 4d reconstruction from a single video.
\newblock 2024{\natexlab{a}}.

\bibitem[Wang et~al.(2024{\natexlab{b}})Wang, Wei, Lu, and Kang]{wang2024plgs}
Yu Wang, Xiaobao Wei, Ming Lu, and Guoliang Kang.
\newblock Plgs: Robust panoptic lifting with 3d gaussian splatting.
\newblock \emph{arXiv preprint arXiv:2410.17505}, 2024{\natexlab{b}}.

\bibitem[Wu et~al.(2024)Wu, Meng, Li, Wu, Shi, Cheng, Zhao, Feng, Ding, Wang, and Zhang]{wu2024opengaussian}
Yanmin Wu, Jiarui Meng, Haijie Li, Chenming Wu, Yahao Shi, Xinhua Cheng, Chen Zhao, Haocheng Feng, Errui Ding, Jingdong Wang, and Jian Zhang.
\newblock Opengaussian: Towards point-level 3d gaussian-based open vocabulary understanding.
\newblock 2024.

\bibitem[Xiang et~al.(2024)Xiang, Li, Lai, Zhang, Liao, Cheng, and Liu]{xiang2024gaussianroom}
Haodong Xiang, Xinghui Li, Xiansong Lai, Wanting Zhang, Zhichao Liao, Kai Cheng, and Xueping Liu.
\newblock Gaussianroom: Improving 3d gaussian splatting with sdf guidance and monocular cues for indoor scene reconstruction.
\newblock \emph{arXiv preprint arXiv:2405.19671}, 2024.

\bibitem[Yan et~al.(2024{\natexlab{a}})Yan, Qu, Xu, Zhao, Wang, Wang, and Li]{yan2024gs_slam}
Chi Yan, Delin Qu, Dan Xu, Bin Zhao, Zhigang Wang, Dong Wang, and Xuelong Li.
\newblock {GS-SLAM}: Dense visual slam with 3d gaussian splatting.
\newblock In \emph{IEEE/CVF Conference on Computer Vision and Pattern Recognition}, pages 19595--19604, 2024{\natexlab{a}}.

\bibitem[Yan et~al.(2024{\natexlab{b}})Yan, Zhang, Zhu, and Wang]{yan2024maskclustering}
Mi Yan, Jiazhao Zhang, Yan Zhu, and He Wang.
\newblock Maskclustering: View consensus based mask graph clustering for open-vocabulary 3d instance segmentation.
\newblock In \emph{IEEE/CVF Conference on Computer Vision and Pattern Recognition}, pages 28274--28284, 2024{\natexlab{b}}.

\bibitem[Yang et~al.(2022)Yang, Li, Zhai, Ming, Liu, and Zhang]{vox-fusion}
Xingrui Yang, Hai Li, Hongjia Zhai, Yuhang Ming, Yuqian Liu, and Guofeng Zhang.
\newblock {Vox-Fusion}: Dense tracking and mapping with voxel-based neural implicit representation.
\newblock In \emph{IEEE International Symposium on Mixed and Augmented Reality}, pages 499--507, 2022.

\bibitem[Yang et~al.(2024)Yang, Gao, Zhou, Jiao, Zhang, and Jin]{yang2024deformable_gaussian}
Ziyi Yang, Xinyu Gao, Wen Zhou, Shaohui Jiao, Yuqing Zhang, and Xiaogang Jin.
\newblock Deformable {3D} gaussians for high-fidelity monocular dynamic scene reconstruction.
\newblock In \emph{IEEE/CVF Conference on Computer Vision and Pattern Recognition}, pages 20331--20341, 2024.

\bibitem[Ye et~al.(2024)Ye, Danelljan, Yu, and Ke]{gaussian_grouping}
Mingqiao Ye, Martin Danelljan, Fisher Yu, and Lei Ke.
\newblock Gaussian grouping: Segment and edit anything in 3d scenes.
\newblock In \emph{European Conference on Computer Vision}, 2024.

\bibitem[Yi et~al.(2024)Yi, Fang, Wang, Wu, Xie, Zhang, Liu, Tian, and Wang]{yi2023gaussiandreamer}
Taoran Yi, Jiemin Fang, Junjie Wang, Guanjun Wu, Lingxi Xie, Xiaopeng Zhang, Wenyu Liu, Qi Tian, and Xinggang Wang.
\newblock Gaussiandreamer: Fast generation from text to 3d gaussians by bridging 2d and 3d diffusion models.
\newblock In \emph{IEEE/CVF Conference on Computer Vision and Pattern Recognition}, 2024.

\bibitem[Yin et~al.(2024)Yin, Liu, Xiao, Cohen-Or, Huang, and Chen]{yin2024sai3d}
Yingda Yin, Yuzheng Liu, Yang Xiao, Daniel Cohen-Or, Jingwei Huang, and Baoquan Chen.
\newblock Sai3d: Segment any instance in 3d scenes.
\newblock In \emph{IEEE/CVF Conference on Computer Vision and Pattern Recognition}, pages 3292--3302, 2024.

\bibitem[Yu et~al.(2024)Yu, Sattler, and Geiger]{yu2024gof}
Zehao Yu, Torsten Sattler, and Andreas Geiger.
\newblock Gaussian opacity fields: Efficient adaptive surface reconstruction in unbounded scenes.
\newblock \emph{ACM Transactions on Graphics}, 2024.

\bibitem[Zhai et~al.(2024)Zhai, Huang, Hu, Li, Bao, and Zhang]{nis-slam}
Hongjia Zhai, Gan Huang, Qirui Hu, Guanglin Li, Hujun Bao, and Guofeng Zhang.
\newblock {NIS-SLAM}: Neural implicit semantic {RGB-D} {SLAM} for {3D} consistent scene understanding.
\newblock \emph{IEEE Transactions on Visualization and Computer Graphics}, 30\penalty0 (11):\penalty0 7129--7139, 2024.

\bibitem[Zhai et~al.(2025{\natexlab{a}})Zhai, Zhang, Zhao, Li, He, Cui, Bao, and Zhang]{splatloc}
Hongjia Zhai, Xiyu Zhang, Boming Zhao, Hai Li, Yijia He, Zhaopeng Cui, Hujun Bao, and Guofeng Zhang.
\newblock {SplatLoc}: 3d gaussian splatting-based visual localization for augmented reality.
\newblock \emph{IEEE Transactions on Visualization and Computer Graphics}, pages 1--11, 2025{\natexlab{a}}.

\bibitem[Zhai et~al.(2025{\natexlab{b}})Zhai, Zhao, Li, Pan, He, Cui, Bao, and Zhang]{neuraloc}
Hongjia Zhai, boming Zhao, Hai Li, Xiaokun Pan, Yijia He, Zhaopeng Cui, Hujun Bao, and Guofeng Zhang.
\newblock Neuraloc: Visual localization in neural implicit map with dual complementary features.
\newblock In \emph{IEEE International Conference on Robotics and Automation}, 2025{\natexlab{b}}.

\bibitem[Zhou et~al.(2024{\natexlab{a}})Zhou, Shao, Xu, Bai, Qiu, Liu, Wang, Geiger, and Liao]{hugs}
Hongyu Zhou, Jiahao Shao, Lu Xu, Dongfeng Bai, Weichao Qiu, Bingbing Liu, Yue Wang, Andreas Geiger, and Yiyi Liao.
\newblock Hugs: Holistic urban 3d scene understanding via gaussian splatting.
\newblock In \emph{IEEE/CVF Conference on Computer Vision and Pattern Recognition}, pages 21336--21345, 2024{\natexlab{a}}.

\bibitem[Zhou et~al.(2024{\natexlab{b}})Zhou, Chang, Jiang, Fan, Zhu, Xu, Chari, You, Wang, and Kadambi]{zhou2024feature_3dgs}
Shijie Zhou, Haoran Chang, Sicheng Jiang, Zhiwen Fan, Zehao Zhu, Dejia Xu, Pradyumna Chari, Suya You, Zhangyang Wang, and Achuta Kadambi.
\newblock Feature 3dgs: Supercharging 3d gaussian splatting to enable distilled feature fields.
\newblock In \emph{IEEE/CVF Conference on Computer Vision and Pattern Recognition}, pages 21676--21685, 2024{\natexlab{b}}.

\bibitem[Zhou et~al.(2021)Zhou, Zhang, and Foroosh]{zhou2021panoptic}
Zixiang Zhou, Yang Zhang, and Hassan Foroosh.
\newblock Panoptic-polarnet: Proposal-free lidar point cloud panoptic segmentation.
\newblock In \emph{IEEE/CVF Conference on Computer Vision and Pattern Recognition}, pages 13194--13203, 2021.

\bibitem[Zhu et~al.(2025)Zhu, Qiu, Wu, Hui, Heng, and Fu]{zhu2025pcf-lifting}
Runsong Zhu, Shi Qiu, Qianyi Wu, Ka-Hei Hui, Pheng-Ann Heng, and Chi-Wing Fu.
\newblock Pcf-lift: Panoptic lifting by probabilistic contrastive fusion.
\newblock In \emph{European Conference on Computer Vision}, pages 92--108. Springer, 2025.

\bibitem[Zuo et~al.(2024)Zuo, Samangouei, Zhou, Di, and Li]{zuo2024fmgs}
Xingxing Zuo, Pouya Samangouei, Yunwen Zhou, Yan Di, and Mingyang Li.
\newblock {FMGS}: Foundation model embedded {3D} gaussian splatting for holistic {3D} scene understanding.
\newblock \emph{International Journal of Computer Vision}, pages 1--17, 2024.

\end{thebibliography}

}


\end{document}